\documentclass[runningheads]{llncs}

\usepackage[T1]{fontenc}
\usepackage{newtxtext}       %
\usepackage[varvw]{newtxmath}
\usepackage{cite}
\usepackage{todonotes}
\usepackage{rotating}
\usepackage{rotating}
\usepackage{makecell}
\usepackage{url}
\usepackage{multirow}
\usepackage{multicol}
\usepackage{tikz}
\usepackage{graphicx}
\usepackage{booktabs}
\usepackage{pifont}
\usepackage{amsmath}
\usepackage[table]{xcolor}
\usepackage{colortbl}
\newcommand{\cmark}{\ding{51}}
\newcommand{\xmark}{\ding{55}}
\usepackage{arydshln}
\begin{document}
\title{Realistic Robustness Evaluation of Tabular Data Driven Models}

\author{Miguel Silva\inst{1}\orcidID{0009-0008-6630-9939} \and
Eva Maia\inst{1}\orcidID{0000-0002-8075-531X} \and
João Vitorino\inst{1}\orcidID{0000-0002-4968-3653} \and
Isabel Praça\inst{1}\orcidID{0000-0002-2519-9859}}

\authorrunning{M. Silva et al.}

\institute{GECAD, ISEP, Polytechnic of Porto, rua Dr. António Bernardino de Almeida, 4249-015 Porto, Portugal
\email{mdgsa@isep.ipp.pt}
}
\maketitle              
\begin{abstract}
The increasing adoption of Machine Learning (ML) models in high-stakes applications has raised critical concerns regarding their robustness against adversarial attacks. While extensive research has focused on adversarial example generation in image and text domains, the study of such examples in tabular data remains comparatively limited, despite its widespread use in real-world systems. This work introduces AURORA, a modular, open-source tool for evaluating the robustness of ML models on tabular data classification. It integrates multiple adversarial perturbation methods and evaluation metrics within a unified framework to systematically assess model behavior under adversarial conditions. Central to AURORA is its evaluation report, which focuses on robustness analysis by quantifying performance degradation, comparing the impact of different perturbation strategies, and identifying model vulnerabilities across features and input regions. The report is complemented by visual artifacts that facilitate the interpretability of robustness patterns. Additionally, AURORA incorporates a distance adjustment mechanism that penalizes implausible adversarial samples based on feature-level constraints, ensuring that the robustness assessment remains grounded in realistic perturbations and is therefore more reliable.


\end{abstract}

\section{Introduction}

The rapid growth of Artificial Intelligence (AI) and its widespread integration into everyday applications has positioned it as a one-size-fits-all transformative technology across virtually every domain. However, this integration can introduce inherent vulnerabilities of AI models, particularly concerning privacy, fairness, and system security~\cite{10.1145/3624010}. These concerns become especially critical in high-stakes environments, such as autonomous vehicles or automated surgical systems, where AI failures can cause unintended and potentially irreversible harm.

Adversarial examples, a term coined by Szegedy et al.~\cite{szegedy2014intriguingpropertiesneuralnetworks}, are inputs that have been intentionally perturbed to cause an AI model to misclassify them, while remaining indistinguishable from the original to the human eye. The vulnerability of models to such subtle input modifications has motivated extensive research across multiple domains. In computer vision, adversarial examples can be crafted by suppressing imperceptible details such as color depth, degrading overall model performance~\cite{Duan_2021_ICCV}, or by modifying as little as a single pixel~\cite{9534332}. In the context of autonomous driving, studies have shown that minor physical alterations to traffic signs, such as the addition of a sticker, can significantly reduce a model's confidence or cause it to misidentify the sign entirely~\cite{etim2025adversarial, HUANG2021102120}.

When adversarial examples are deliberately created to cause a model to misbehave, this is referred to as an adversarial attack. Such attacks have been extensively studied for other domains, such as text classifiers, although their construction differs fundamentally from image-based attacks. In this setting, the objective is to alter the model's prediction by modifying parts of the input text, while preserving its readability. Such modifications can be introduced at the character, word, sentence, or multi-level~\cite{QIU2022278}, and may include deliberate misspellings~\cite{pruthi-etal-2019-combating}, among other strategies.

In tabular data, however, comparatively little work has addressed the constraints inherent to this setting. Unlike images or text, tabular inputs are highly structured, with each column corresponding to a specific attribute. Adversarial perturbation methods developed for images are therefore not directly applicable to tabular data, as they fail to account for its structure~\cite{HE2025200461}, and the similarity metrics used to constrain image perturbations are not transferable to this context. Furthermore, no universal robustness metric exists for measuring model resilience against adversarial attacks, and existing approaches vary considerably in scope and methodology, as some focus on overall model performance based on predicted labels~\cite{10.1145/3374664.3375728}, while others rely on metrics such as the divergence between adversarial and true class predictions~\cite{9325360}. These gaps are more evident in tabular settings, where the absence of domain-specific constraints and evaluation standards makes it difficult to assess the robustness of models.

To address these limitations, this work presents AURORA, a tool for evaluating the robustness of Machine Learning (ML) models to adversarial attacks on tabular data. By integrating diverse perturbation methods, evaluation metrics, and a distance adjustment mechanism, AURORA ensures that the adversarial examples used for evaluation are both realistic and valid, resulting in a more reliable and meaningful robustness assessment. In summary, the main contributions of this work are:
\begin{itemize}
\item presentation of AURORA, an open-source framework for the systematic evaluation of ML model robustness on tabular data under adversarial conditions;
\item introduction of a distance adjustment mechanism that ensures a more reliable robustness evaluation by accounting for the quality of adversarial examples;
\item validation of AURORA through a case study evaluating a CatBoost model trained for network intrusion detection.
\end{itemize}

This paper is organized as follows. Section~\ref{AURORA} introduces AURORA and provides an overview of each perturbation method, evaluation metric, and distance calculation. Section~\ref{related_work} reviews existing tools for generating adversarial examples and evaluating model robustness, discussing their limitations in the context of tabular data and positioning AURORA in relation to them. Section~\ref{eval} describes the experimental setup and presents and discusses the robustness evaluation results for a tree-based ML model. Finally, Section~\ref{conclusion} summarizes the main conclusions and outlines directions for future work.

\section{Related Work}
\label{related_work}

ML robustness centers on evaluating how a model's predictions behave when inputs are presented with subtle perturbations. Extensive research has focused on developing several methodologies to test models by subjecting them to multiple attack vectors and quantifying performance degradation \cite{10.1145/3743126}. Following this line of research, various studies have employed different metrics and perturbation methods to assess this vulnerability~\cite{JIN2023118936,9774934,10.1145/3664647.3681676}. This necessity led to the development of comprehensive toolboxes, including the Adversarial Robustness Toolbox (ART)~\cite{art2018}, CleverHans~\cite{papernot2018cleverhans}, Foolbox~\cite{rauber2017foolbox,rauber2017foolboxnative}, AdvBox~\cite{goodman2020advbox}, Azure Counterfit\footnote{https://github.com/Azure/counterfit}, and TabularBench~\cite{simonetto2024tabularbench}. However, these tools exhibit some limitations. While they provide the necessary functionality for generating adversarial examples, most are designed primarily for the generation of attacks rather than for systematic comparative evaluation. For instance, although ART provides advanced metrics such as CLEVER and loss sensitivity, using these robust evaluation metrics is optional, placing a significant interpretive burden on the user to select the appropriate metrics. Furthermore, other specialized frameworks lack a standardized mechanism for comparing the efficacy or severity of multiple perturbation methods against a model.

Beyond the application of a tool, evaluating ML robustness requires careful consideration of the underlying data type and the perturbation methods employed. While considerable research has been conducted focusing on computer-vision models, much less work has been done to evaluate ML robustness in tabular data. In image-based settings, perturbation methods such as Carlini \& Wagner, ZerothOrderOptimization, BoundaryAttack, and the Fast Gradient Sign Method \cite{goodfellow2015explainingharnessingadversarialexamples} often operate under the assumption of minimal perturbation defined by $l_p$-norms ($\ell_1$, $\ell_2$, or $\ell_\infty$)~\cite{lp,10643066,lp1}. This constraint allows researchers to manage a trade-off between imperceptibility and attack success rate~\cite{10.1145/3510548.3519370}. However, this established methodology does not translate reliably to tabular data, as perturbations created via these standard $l_p$-norms frequently violate the inherent structural constraints or statistical properties of real-world tabular features~\cite{HE2025200461}. Consequently, evaluation metrics in such settings often assume high sample quality by assessing model performance without validating individual sample realism \cite{10624973,10488755}, even when these methods generate invalid examples that are nevertheless used for scoring~\cite{BARIK2024100681,10932405}.

Overall, these limitations underscore two significant gaps in the literature. First, there is a lack of standardization regarding evaluation metrics. Currently, while tools exist for attack generation, they do not automatically standardize or simplify the calculation into a single, universally interpretable metric. Second, applying perturbation methods developed for computer vision directly to tabular data violates the inherent structural constraints of the domain. This inconsistency highlights the need for a higher level of rigour, namely, a way to evaluate robustness while considering whether an adversarial example is both valid and realistic. To bridge these critical gaps, we propose AURORA. Our tool provides a fully automated evaluation pipeline, calculating a single, standardized robustness score across diverse attack methods and model requirements to be considered robust. Most significantly, AURORA introduces a novel distance adjustment mechanism to distinguish between high or low-quality adversarial examples and penalize the final score accordingly, to ensure a more reliable assessment of model robustness.

\section{AURORA}
\label{AURORA}

AURORA~\footnote{\url{https://github.com/msilva2002/AURORA}} (Adversarial Understanding of Robustness and Offensive Resilience Analysis) is an open-source tool that enables users to assess the robustness of ML models. This assessment is done by subjecting clean data and the model to multiple perturbation methods in order to create adversarial examples specific to the model. Each method perturbs clean samples according to its own objective and strategy, producing adversarial examples of varying characteristics for the same input dataset. The resulting adversarial sets are evaluated independently, allowing users to identify which methods are most efficient and most effective against their model.

The evaluation is conducted using a diverse set of metrics, each targeting a different aspect of robustness, ensuring a comprehensive yet interpretable assessment. To further improve reliability and ensure the evaluation is based on realistic perturbations, AURORA applies a distance adjustment mechanism to key metrics. This mechanism penalizes results derived from adversarial examples that are either invalid, violating the structural constraints of the feature space, such as implausible categorical transitions, or of low quality, where the perturbation distance exceeds the defined plausibility threshold. As such, the calculated distance directly grounds the resulting robustness score, ensuring that the final evaluation reflects only realistic and feasible attack scenarios. Moreover, since this distance is computed relative to the clean samples, it also ensures a domain-specific evaluation, aligned with the actual characteristics of the dataset.

Once all perturbation methods have been evaluated, AURORA consolidates the results in a report addressing two robustness scenarios. The standard scenario aggregates performance across all perturbation methods, providing a broad assessment of model robustness under typical adversarial conditions. The worst-case scenario considers only the most effective perturbation method for each attack objective, offering a more conservative and critical evaluation suited for high-stakes applications where the strongest possible adversary must be assumed.

\subsection{Perturbation Methods}

AURORA is designed to provide a structured and reliable evaluation of the robustness of ML models under adversarial conditions in tabular domains. It explicitly accounts for the limitations of applying generic perturbation methods to tabular data, where feature values lack perceptual structure and even small modifications may violate meaningful domain constraints while remaining difficult to detect for an analyst.
To address this, AURORA embeds adversarial generation directly within a controlled evaluation pipeline. It enforces feature-level constraints during the construction of adversarial examples, ensuring that generated samples remain consistent with valid data ranges and dataset-specific rules. In addition, it incorporates a distance adjustment mechanism that penalizes invalid or unrealistic perturbations when computing robustness metrics. This guarantees that the evaluation is not biased by implausible samples and instead reflects the model’s behavior under realistic adversarial settings.

To guarantee consistency and comparability across methods, AURORA defines a set of requirements for all integrated perturbation techniques: each method must generate an adversarial attempt for every input sample, preserve the original sample ordering to enable accurate distance computations, respect logical and dataset-specific feature constraints, and support targeted attack variants. Additionally, AURORA prioritizes the inclusion of black-box methods, allowing robustness evaluation without requiring access to model internals.

Based on these criteria, AURORA incorporates five widely used perturbation methods (Table~\ref{tab:adversarial_methods}), each implemented in both targeted and untargeted variants. With the exception of the Carlini \& Wagner method, which requires gradient access, all methods operate in a black-box setting and rely solely on model queries, using predicted labels or confidence scores. When the number of classes is not explicitly available, AURORA automatically infers it by querying the model across the dataset.

\begin{table}[htpb]
\centering
\caption{Adversarial methods implemented in AURORA}
\label{tab:adversarial_methods}
\begin{tabular}{@{}ccccc@{}}
\toprule
Perturbation         & Is & Requires & Requires \\Method&Targeted&Knowledge&Constraints \\ \midrule
A2PM \cite{a2pm}           &     \cmark/\xmark    &   \tikz\draw[fill=black] (0,0) circle (0.15cm);                 &      -       \\
BoundaryAttack \cite{boundary} &     \cmark/\xmark    &   \tikz\draw[fill=black] (0,0) circle (0.15cm);                 &      \cmark       \\
Carlini \& Wagner \cite{cw}  &     \cmark/\xmark    &   \tikz\draw[fill=white] (0,0) circle (0.15cm);                 &      \cmark       \\
HopSkipJumpAttack \cite{hopskipjump}    &     \cmark/\xmark    &   \tikz\draw[fill=black] (0,0) circle (0.15cm);                 &      \cmark       \\
Zeroth Order Optimization \cite{zoo}            &     \cmark/\xmark    &   \tikz\draw[fill=black] (0,0) circle (0.15cm);                 &      \cmark       \\ \bottomrule
\end{tabular}
\end{table}

Furthermore, given that most existing perturbation methods were originally designed for other domains, AURORA provides both unconstrained and constrained implementations. In the constrained versions, a validation step enforces feature-level rules after perturbation: float values are rounded to integers where appropriate, and for categorical features, only the highest value is kept as active, as one, while all others are set to zero. This ensures that generated samples remain within valid ranges and respect dataset-specific constraints. The dual implementation enables direct comparison between unconstrained and constraint-aware attacks, highlighting the impact of validity on robustness evaluation.

\subsection{Evaluation Metrics}

To evaluate the resulting perturbations, AURORA employs several widely used metrics.

\textbf{Clean Accuracy} (CA), measures the performance of the model on unperturbed data. It reflects the model’s ability to generalize to unseen inputs and serves as a baseline for comparison with adversarial performance. This metric requires access to ground-truth labels.

\textbf{Adversarial Accuracy} (AA) evaluates model performance on adversarially perturbed samples. It provides a direct comparison to Clean Accuracy, quantifying the degradation in predictive performance caused by adversarial manipulation. This metric also requires ground-truth labels.

\textbf{Misclassification Rate}  (MR) measures the proportion of predictions that differ between clean and adversarial samples. Unlike accuracy-based metrics, it does not require ground-truth labels, making it suitable for black-box evaluation scenarios. However, it does not indicate whether the new prediction is correct.

\textbf{Attack Success Rate} (ASR), quantifies the effectiveness of targeted adversarial attacks. It measures the proportion of adversarial samples that are successfully classified into a specific target class defined by the attacker~\cite{asr1}. This metric is widely used under different names, such as fooling ratio or evasion rate.

\textbf{Attack Deterioration} (AD) measures the relative degradation in model performance caused by adversarial perturbations. By comparing Clean Accuracy and Adversarial Accuracy, it provides a normalized measure of robustness loss~\cite{ad1,ad}.

\textbf{Confusion Matrix} provides a detailed breakdown of prediction changes. By treating clean predictions as reference labels and adversarial predictions as outputs, it enables the analysis of class-specific vulnerabilities and transition patterns.

Finally, \textbf{Time Required} measures the computational cost of each perturbation method. While not directly related to attack effectiveness, it provides practical insights into the feasibility of applying each method in real-world scenarios.

\subsection{Distance Adjustment Mechanism}

AURORA introduces a distance adjustment mechanism to quantify the deviation between adversarial and clean samples, with the goal of assessing the realism of generated perturbations. This mechanism combines the Euclidean distance between clean and adversarial examples, scaled for numerical features, and the Hamming distance for categorical features, enabling a unified treatment of heterogeneous tabular data.

Building on this distance measure, AURORA incorporates a penalty scheme that down-weights invalid or implausible adversarial examples. These penalties explicitly distinguish realistic perturbations from those that violate feature constraints and would not occur in practice. An adversarial sample is considered invalid if:

\begin{itemize}
    \item categorical features containing values outside the valid range (not equal to 0 or 1);
    \item multiple active values within a single categorical feature (violating one-hot encoding);
    \item absence of an active value within a categorical feature. Although this can occur, if a feature is omitted from the input columns, then all of its values will be set to 0. Since perturbation methods cannot distinguish between a true zero and a zero that is the result of an omitted feature, such cases are considered invalid.
\end{itemize}

To ensure that robustness evaluation reflects only realistic perturbations, AURORA defines a dataset-specific threshold over the computed distance, representing the maximum acceptable deviation for a perturbation to be considered plausible. This threshold is derived from the characteristics of successfully perturbed samples, incorporating statistics such as the median absolute values of numerical features and possible valid categorical transitions. Since categorical features are one-hot encoded, a valid transition corresponds to a Hamming distance of 2, as flipping one category to another changes exactly two positions. For numerical features, the median is used rather than the mean, as it is computed solely from the clean dataset and is less sensitive to skewed or non-linear feature distributions. Its formulation is given in Equation~\eqref{eqn:threshold}, where $\mathcal{C}$ is the number of categorical features, $\mathcal{N}$ is the set of numerical features, and $X_j$ represents the clean samples' values for numerical feature $j$.

\begin{equation}
\text{Threshold} = 2\mathcal{C} + \sum_{j \in \mathcal{N}} \left| \text{median}(X_j) \right|
\label{eqn:threshold}
\end{equation}

The computed distance and threshold are then incorporated into AURORA’s distance adjustment mechanism, which reduces the influence of unrealistic perturbations on the evaluation. While prior approaches typically apply evaluation metrics directly to adversarial samples, AURORA adjusts metric contributions based on perturbation quality, ensuring that adversarial examples generated by less suitable methods do not disproportionately affect the results.
The adjustment mechanism is then designed to satisfy the following properties:
\begin{itemize}
    \item the adjusted metric remains non-zero whenever the original metric is non-zero, preserving meaningful performance signals;
    \item a dataset-dependent threshold distance must be defined to characterize the maximum acceptable perturbation magnitude, ensuring that valid and realistic perturbations retain their full contribution to the final metric;
    \item the adjustment introduces a gradual degradation as distance increases, allowing differentiation between mild and severe perturbations; 
    \item the adjustment must strictly increase with distance to ensure that it acts exclusively as a penalty mechanism and never artificially inflates performance metrics.
\end{itemize}

When the distance between a clean and an adversarial sample exceeds the computed threshold, a penalty, with a value of a tenth of the defined threshold for each categorical feature, as shown in Equation~\eqref{eqn:penalty}, is added to the distance measured for each invalid case detected to reduce the contribution of that sample to the final evaluation metric. As the threshold depends on the success of the perturbation methods, so does the penalty. The adjustment follows a step function to ensure a gradual degradation of metric values, as represented in Equation~\eqref{eqn:da}, where $x$ is defined as the positive difference between the perturbation distance and the threshold, taking the value zero whenever this difference is negative. This margin allows room for perturbations to be generated within an acceptable range; without it, perturbations would never be fully accounted for, as any deviation would be penalized immediately.

\begin{equation}
    \text{Penalty} = \frac{\text{Threshold}}{10} \cdot n_{categorical}
    \label{eqn:penalty}
\end{equation}

\begin{equation}
    \text{Distance Adjustment} =
    \begin{cases}
        1, & \text{if } x = 0 \\
        
        1 - \dfrac{0.9x}{\text{Threshold}}, & \text{if } 0 < x \leq \text{Threshold} \\
        \\
        1 - \bigg(1 - \dfrac{\text{Threshold}}{20\left(x - \dfrac{\text{Threshold}}{2}\right)}\bigg), & \text{if } x > \text{Threshold}
    \end{cases}
    \label{eqn:da}
\end{equation}

Finally, AURORA applies this mechanism to derive distance-adjusted evaluation metrics, including Distance Adjusted MR and Distance Adjusted ASR. These metrics extend their standard counterparts by weighting their contributions according to perturbation quality. The adjusted metric is computed as shown in Equation~\eqref{eqn:da_metric}, where Metric denotes the original, unadjusted value.
\begin{equation}
    \text{Distance Adjusted Metric} = \text{Metric} * \text{Distance Adjustment}
    \label{eqn:da_metric}
\end{equation}
Although this adjustment can, in principle, be applied to any metric defined over adversarial samples, it is not directly suitable for composite metrics that jointly depend on clean and adversarial data, such as Attack Deterioration, where distance-based penalties would confound the contributions of the underlying components.

\subsection{Robustness Metric}

AURORA introduces a composite robustness score that consolidates multiple complementary metrics into a single, unified, and interpretable measure of adversarial robustness. Rather than relying on isolated metrics that capture only a specific aspect of a model's robustness, this score integrates three core metrics that collectively cover the primary objectives of adversarial attacks: the Distance Adjusted Attack Success Rate (DAASR), which measures the adversary's ability to force targeted misclassification; the Distance Adjusted Misclassification Rate (DAMR), which captures general prediction instability induced by perturbations; and Attack Deterioration, which quantifies the overall degradation of model performance, as shown in Equation~\eqref{eqn:robustness}. By combining these three dimensions, the resulting score is expressed on a normalized scale from 0 to 100 and partitioned into five interpretable robustness levels: \textit{Not Robust} (0-20), \textit{Weakly Robust} (21-40), \textit{Moderately Robust} (41-60), \textit{Robust} (61-80), and \textit{Very Robust} (81-100).

\begin{equation}
    \text{Robustness score} = 100 - \frac{\text{DAASR} + \text{DAMR} + \text{Attack Deterioration}}{3}
    \label{eqn:robustness}
\end{equation}

To account for the diversity of deployment contexts, AURORA further instantiates the robustness score under two distinct evaluation scenarios. In the first scenario, a standard evaluation is made, where the robustness score of the model results from the average performance computed across all perturbation methods, regardless of their individual results. While this approach can undermine high-performing perturbation methods, it offers a general assessment of model robustness under typical adversarial conditions and provides insight into how the model performs when subjected to a variety of perturbation strategies.

The second scenario, by contrast, is the worst-case scenario. It selects the most effective perturbation method for each attack objective and always results in a lower evaluation, or at most an equal one, than the first scenario. Although this approach does not represent typical performance, it provides a pessimistic view of model robustness by considering the maximum adversarial threat. This scenario is more relevant for high-risk applications, where the model may encounter the strongest possible adversary and failures may have significant consequences.

\section{Experimental Evaluation}
\label{eval}

A case study was conducted to demonstrate AURORA's capability to assess ML model robustness in a real-world tabular setting. This section describes the experimental setup, including the dataset, feature selection, and model configuration, followed by a detailed analysis of the report created by AURORA, including robustness results and visual artifacts.

\subsection{Dataset and Models}

To demonstrate AURORA's evaluation capabilities, we assess the robustness of an ML classifier on the GeNIS~\cite{SILVA2025111487}, a tabular dataset designed for network intrusion detection scenarios. 

The high dimensionality of the dataset makes dimensionality reduction a necessary step to make the evaluation computationally feasible without sacrificing discriminative power. To this end, Recursive Feature Elimination was applied, yielding two numerical key features: SrcBytes, the number of bytes in the transaction from source to destination, SIntPktMax, the maximum value of the source interpacket arrival time, and two categorical features, Proto and State, representing protocol types and connection states, respectively. Although Sdaddr was classified as the most important feature, it was removed since it reflects the number of connections sharing the same service and source IP, and its inclusion would undermine generalization to unseen hosts.

Given the predominance of categorical features, a tree-based CatBoost classifier was selected and optimized using 5-fold cross-validation, with the macro-averaged F1-score as the evaluation metric for multi-class classification. The best hyperparameters found were a learning rate of 0.1, a depth of 6, and 1000 iterations. The model achieved near-perfect classification performance on the GeNIS dataset, with 99.74\% accuracy, 98.61\% macro-averaged precision, 97.08\% recall, 97.82\% F1-score, and 2.15\% false positive rate. FPR was defined as the proportion of benign samples incorrectly classified as attacks, that is, samples with a true label of 0 predicted as any other class.

\subsection{Robustness Results}

To evaluate the robustness of the CatBoost model using AURORA, 1000 random samples were extracted from the test set, retaining only the features used during training along with their true labels. Since methods such as A2PM require information about the distribution of the data, gathering a representative set of clean examples is essential to ensure a reliable evaluation. Although 1000 samples were used in this evaluation, other sample sizes may also be used. These samples were sent to AURORA and served as the clean baseline for the perturbation methods. For the objective of AURORA's targeted perturbation method, a file containing the label 0 was used, meaning that samples that are already classified as 0 (263 samples) would be skipped, and the others (737 samples) would be perturbed with the objective of being classified as 0. All resulting adversarial examples, configurations, and the final report created by AURORA in these experiments are publicly available on GitHub\footnote{\url{https://github.com/msilva2002/AURORA-Realistic-Robustness-Evaluation}}.

All perturbation methods were configured with default settings, with the exception of A2PM and the constrained variants. For A2PM, each categorical feature and its combinations were explicitly defined with a probability value of 0.8, and the numerical features, SrcBytes and SIntPktMax, were set with minimum and maximum perturbation ratios of 0.001 and 0.2, respectively, while SrcBytes was set as an integer. For the constrained variants, State and Proto were set as categorical features, and SrcBytes was set as an integer. 

Two sets of adversarial results were obtained using AURORA. The first set contained the perturbation methods adapted with constraints alongside A2PM, and the second method the perturbation methods as they were originally developed. The evaluation results for each set of adversarial examples created by each perturbation method are summarized in Table~\ref{tab:catboost_adv}. Since the CatBoost model provides no gradient information, the Carlini \& Wagner method was omitted, as it is incompatible with gradient-free settings.

\begin{table}[!htbp]
\caption{Results obtained for each perturbation method against CatBoost, on top the set with constraints and A2PM, on the bottom the original perturbation methods. BA: BoundaryAttack; HSJA: HopSkipJumpAttack; ZOO: Zeroth Order Optimization; T: targeted; C: constrained; D: distance; TH: threshold; P: penalty; CA: clean accuracy (\%); AA: adversarial accuracy (\%); ASR: attack success rate (\%); MR: misclassification rate (\%); AD: attack deterioration (\%); TR: time required (s); DAASR: distance-adjusted attack success rate (\%); DAMR: distance-adjusted misclassification rate (\%).}
\centering
\label{tab:catboost_adv}
\begin{tabular}{@{}cccccccccccccc@{}}
\toprule
    \textbf{Method} & \textbf{T} & \textbf{C} & \textbf{D} & \textbf{TH} & \textbf{P} & \textbf{CA} & \textbf{AA} & \textbf{ASR} & \textbf{MR} & \textbf{AD} & \textbf{TR} & \makecell{\textbf{DA} \\ \textbf{ASR}} & \makecell{\textbf{DA} \\ \textbf{MR}} \\ \midrule

A2PM & \cellcolor{red!20} \xmark & - & 2.05 & 4.03 & 0.81 & 99.3 & 11.3 & - & 89.2 & 88.62 & 0.17 & - & 89.20 \\
BA & \cellcolor{red!20} \xmark & \cellcolor{green!20} \cmark & 2.0 & 4.07 & 0.81 & 99.3 & 90.1 & - & 9.4 & 9.26 & 421.89 & - & 9.4 \\
HSJA & \cellcolor{red!20} \xmark & \cellcolor{green!20} \cmark & 4.0 & 4.05 & 0.81 & 99.3 & 57.3 & - & 42.9 & 42.3 & 196.44 & - & 42.9 \\
ZOO & \cellcolor{red!20} \xmark & \cellcolor{green!20} \cmark & 0.0 & 0.0 & 0.0 & 99.3 & 99.3 & - & 0.0 & 0.0 & 2.24 & - & 0.0 \\
\hdashline
A2PM & \cellcolor{green!20} \cmark & - & 2.04 & 4.02 & 0.8 & 99.3 & 27.8 & 84.53 & 72.1 & 72.0 & 0.16 & 84.53 & 72.10 \\
BA & \cellcolor{green!20} \cmark & \cellcolor{green!20} \cmark & 2.0 & 4.03 & 0.81 & 99.3 & 88.6 & 12.21 & 10.9 & 10.78 & 384.82 & 12.21 & 10.9 \\
HSJA & \cellcolor{green!20} \cmark & \cellcolor{green!20} \cmark & 4.0 & 4.03 & 0.81 & 99.3 & 58.8 & 48.3 & 41.4 & 40.79 & 162.92 & 48.3 & 41.4 \\
ZOO & \cellcolor{green!20} \cmark & \cellcolor{green!20} \cmark & 0.0 & 0.0 & 0.0 & 99.3 & 99.3 & - & 0.0 & 0.0 & 4.5 & - & 0.0 \\
\hline

BA & \cellcolor{red!20} \xmark & \cellcolor{red!20} \xmark & 147.22 & 4.03 & 0.81 & 99.3 & 31.2 & - & 68.8 & 68.58 & 420.11 & - & 0.1 \\
HSJA & \cellcolor{red!20} \xmark & \cellcolor{red!20} \xmark & 147.22 & 4.03 & 0.81 & 99.3 & 18.8 & - & 81.8 & 81.07 & 192.63 & - & 0.12 \\
ZOO & \cellcolor{red!20} \xmark & \cellcolor{red!20} \xmark & 0.0 & 0.0 & 0.0 & 99.3 & 99.3 & - & 0.0 & 0.0 & 2.19 & - & 0.0 \\
\hdashline
BA & \cellcolor{green!20} \cmark & \cellcolor{red!20} \xmark & 145.6 & 4.02 & 0.8 & 99.3 & 26.3 & 100 & 73.7 & 73.51 & 391.76 & 0.14 & 0.11 \\
HSJA & \cellcolor{green!20} \cmark & \cellcolor{red!20} \xmark & 145.6 & 4.02 & 0.8 & 99.3 & 26.3 & 100 & 73.7 & 73.51 & 161.03 & 0.14 & 0.11 \\
ZOO & \cellcolor{green!20} \cmark & \cellcolor{red!20} \xmark & 0.0 & 0.0 & 0.0 & 99.3 & 99.3 & - & 0.0 & 0.0 & 4.3 & - & 0.0 \\\bottomrule
\end{tabular}
\end{table}

As expected, the methods that generated adversarial samples with no regard for tabular data were those with a higher distance. As the threshold is calculated based on the normalized median values of the clean samples that were successfully attacked, this threshold varies only slightly between methods. The best performing untargeted perturbation method was A2PM, as it reached the highest MR, both with the distance adjustment and without. For the targeted version, both BoundaryAttack and HopSkipJumpAttack were able to reach 100\% ASR, but since the perturbed data is neither considered realistic nor valid because the distance is far from the threshold, their distance-adjusted ASR is below 1\%. As such, the best targeted perturbation method that respected the tabular data constraints specified was A2PM. An interesting detail is that the targeted version of BoundaryAttack was able to achieve a higher MR than its untargeted version for both the constrained and unconstrained versions, although in theory, this would not happen, since the target pool is higher for the untargeted attack. Tables~\ref{tab:boundary1} and \ref{tab:boundary2} present the results of this perturbation method. Some adversarial samples were classified as values other than the targeted class 0, since the constraints were applied only after the attack had completed. While this may influence the MR, only the adversarial examples predicted as the target contribute to the ASR. These results suggest that it is easier to create perturbations if the method is aiming for this specific class rather than any other neighboring class, and if class 0 could be perturbed in an untargeted manner, it could reach an even higher MR.

Regarding computational cost, A2PM surpassed by far every other method in terms of efficiency, even surpassing ZerothOrderOptimization, which was unable to create adversarial examples for CatBoost using these samples. Even though the constrained versions require an additional step to ensure the adversarial samples respect the intrinsic properties of tabular data, this did not add much time to the total time required. Interestingly enough, while most perturbation methods took less time to create targeted adversarial examples, as the pool of possible perturbations is lower, ZerothOrderOptimization took longer while providing the same results.

\begin{table}[htpb]
\centering
\begin{minipage}{0.48\textwidth}
\centering
\caption{Confusion matrix of constrained untargeted BoundaryAttack.}
{\setlength{\tabcolsep}{6pt}
\begin{tabular}{cc|cccc|}
\cline{3-6}
 &  & \multicolumn{4}{c|}{Predicted} \\ \cline{3-6}
 &  & \multicolumn{1}{c|}{0} & \multicolumn{1}{c|}{1} & \multicolumn{1}{c|}{2} & 3 \\ \cline{1-6}
\multicolumn{1}{|c|}{\multirow{4}{*}{Actual}} & 0 & \textbf{249} & 6 & 8 & 0 \\ \cline{2-2}
\multicolumn{1}{|c|}{} & 1 & 1 & \textbf{245} & 14 & 0 \\ \cline{2-2}
\multicolumn{1}{|c|}{} & 2 & 48 & 1 & \textbf{378} & 1 \\ \cline{2-2}
\multicolumn{1}{|c|}{} & 3 & 10 & 0 & 5 & \textbf{34} \\ \cline{1-6}
\end{tabular}}
\label{tab:boundary1}
\end{minipage}
\hfill
\begin{minipage}{0.48\textwidth}
\centering
\caption{Confusion matrix of constrained untargeted BoundaryAttack.}
{\setlength{\tabcolsep}{6pt}
\begin{tabular}{cc|cccc|}
\cline{3-6}
 &  & \multicolumn{4}{c|}{Predicted} \\ \cline{3-6}
 &  & \multicolumn{1}{c|}{0} & \multicolumn{1}{c|}{1} & \multicolumn{1}{c|}{2} & 3 \\ \cline{1-6}
\multicolumn{1}{|c|}{\multirow{4}{*}{Actual}} & 0 & \textbf{263} & 0 & 0 & 0 \\ \cline{2-2}
\multicolumn{1}{|c|}{} & 1 & 33 & \textbf{211} & 16 & 0 \\ \cline{2-2}
\multicolumn{1}{|c|}{} & 2 & 49 & 0 & \textbf{379} & 0 \\ \cline{2-2}
\multicolumn{1}{|c|}{} & 3 & 8 & 0 & 3 & \textbf{38} \\ \cline{1-6}
\end{tabular}}
\label{tab:boundary2}
\end{minipage}
\end{table}

As BoundaryAttack operates by pushing clean samples toward the decision boundary of a model, its generated adversarial examples tend to remain closer to the original clean samples than those produced by HopSkipJumpAttack, as can be seen in the distance between these two constrained methods. While HopSkipJumpAttack did create around four times the number of adversarial examples that BoundaryAttack created, as the distance is computed based on the median, HopSkipJumpAttack having a higher distance translates to a farther adversarial example from the original value. For the unconstrained methods, however, this does not happen, as their distance is the same. This is likely due to the number of adversarial samples created being so high that the median converged to the same value.

\begin{figure}[htpb]
    \centering
    \includegraphics[width=\textwidth]{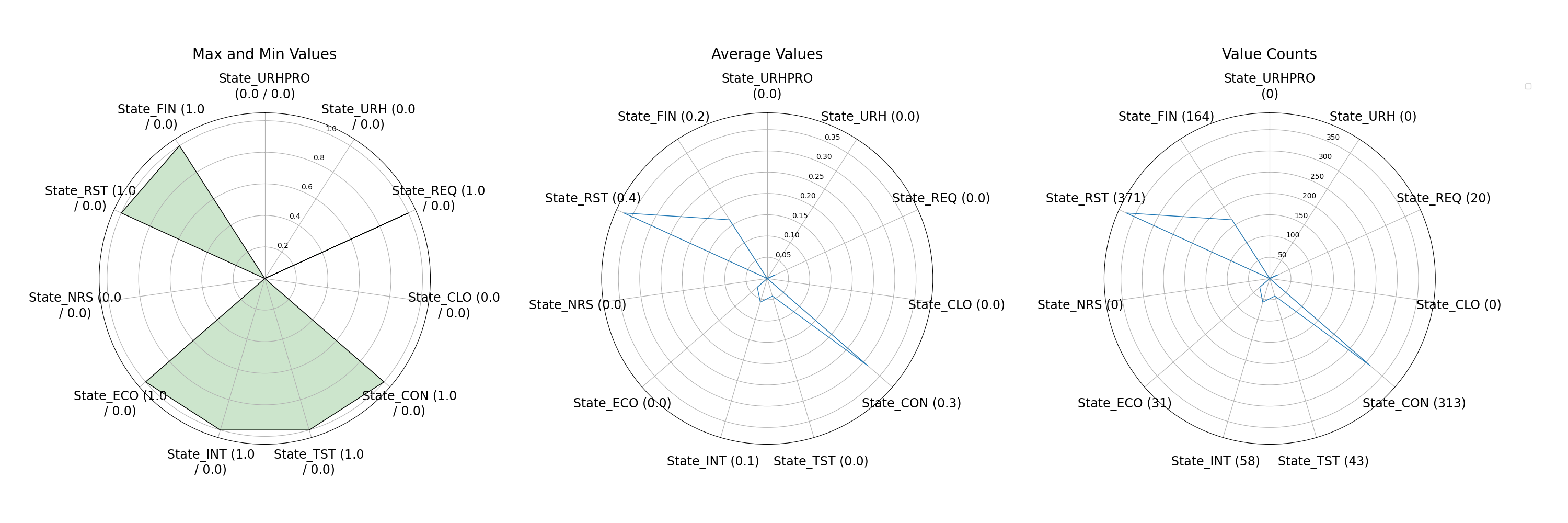}
    \caption{Original State category statistics.}
    \label{fig:state_original}
    \end{figure}

Regarding the modified feature values, and more specifically the categorical feature State, the transactional state of the network flow, the initial statistical distribution of the selected clean samples is shown in Fig.~\ref{fig:state_original}, created by AURORA. In these samples, values such as CLO (State\_CLO) and URH (State\_URH) are absent, since these columns exist, but none of the selected samples have them as values, as seen in "Max and Min Values" diagram in Fig.~\ref{fig:state_original}. The average value of a valid categorical feature should always sum to 1, since each sample can only hold a single categorical value. Consequently, the sum of value counts across all values within the same categorical feature must always equal the number of samples, in this case 1000. Fig.~\ref{fig:state_original} confirms this property for the clean samples, all of which satisfy this constraint.

\begin{figure}[htpb]
    \centering
    \includegraphics[width=\textwidth]{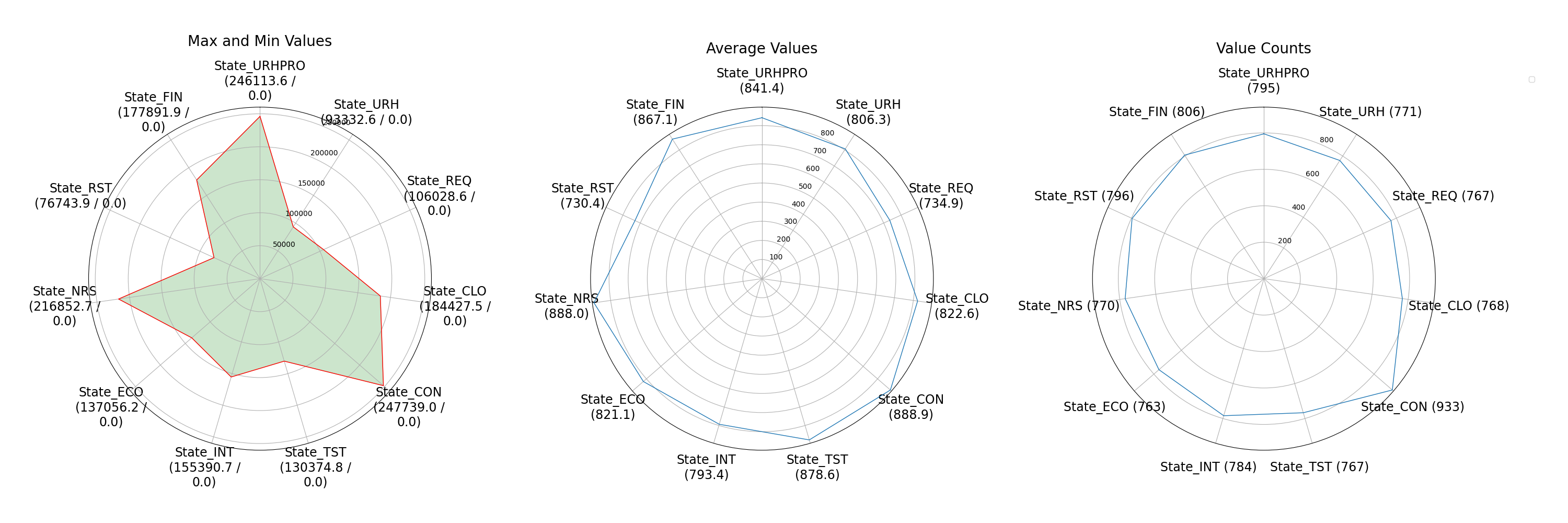}
    \caption{Resulting State category statistics for HopSkipJumpAttack with no constraints.}
    \label{fig:state_hopskipjump}
    \end{figure}

When subjected to an unconstrained method designed for image data, such as HopSkipJumpAttack, the resulting feature distribution differs substantially from the original. As shown in Fig.~\ref{fig:state_hopskipjump}, the adversarial samples exhibit clearly invalid values, with maximum values reaching up to 247739.0 across all values in the State categorical feature. The average values are equally anomalous, with every possible category averaging over 700, which is invalid, since the average is always 1 with valid samples. The "Value Counts" diagram distribution does not retain a similar shape to the "Average Values", and having approximately 800 counts per category also indicates invalid data. Furthermore, the total sum of value counts exceeds 8000, far beyond the expected 1000, a discrepancy that would only be consistent with a dataset of over 8000 samples.

\begin{figure}[htpb]
    \centering
    \includegraphics[scale=0.15]{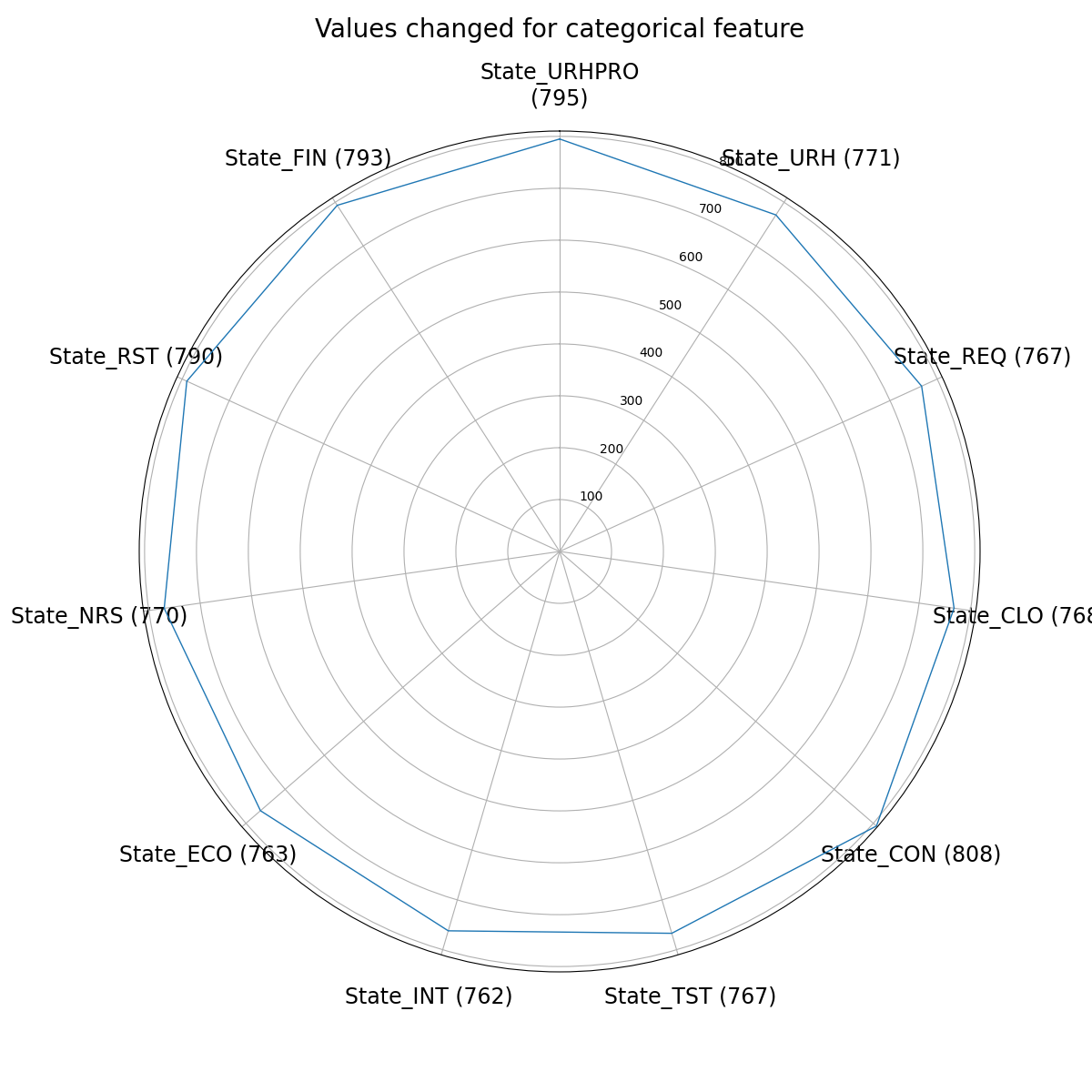}
    \caption{Changes in the State category for HopSkipJumpAttack with no constraints.}
    \label{fig:state_hopskipjump_changes}
    \end{figure}

Each change made to the values of the State categorical feature in each sample can be seen in Fig.~\ref{fig:state_hopskipjump_changes}. As shown, every value underwent an infeasible number of changes, which is only possible under invalid perturbations, given that a set of 1000 samples can produce at most 2000 changes, corresponding to each sample changing from one value to another.

    \begin{figure}[htpb]
    \centering
    \includegraphics[width=\textwidth]{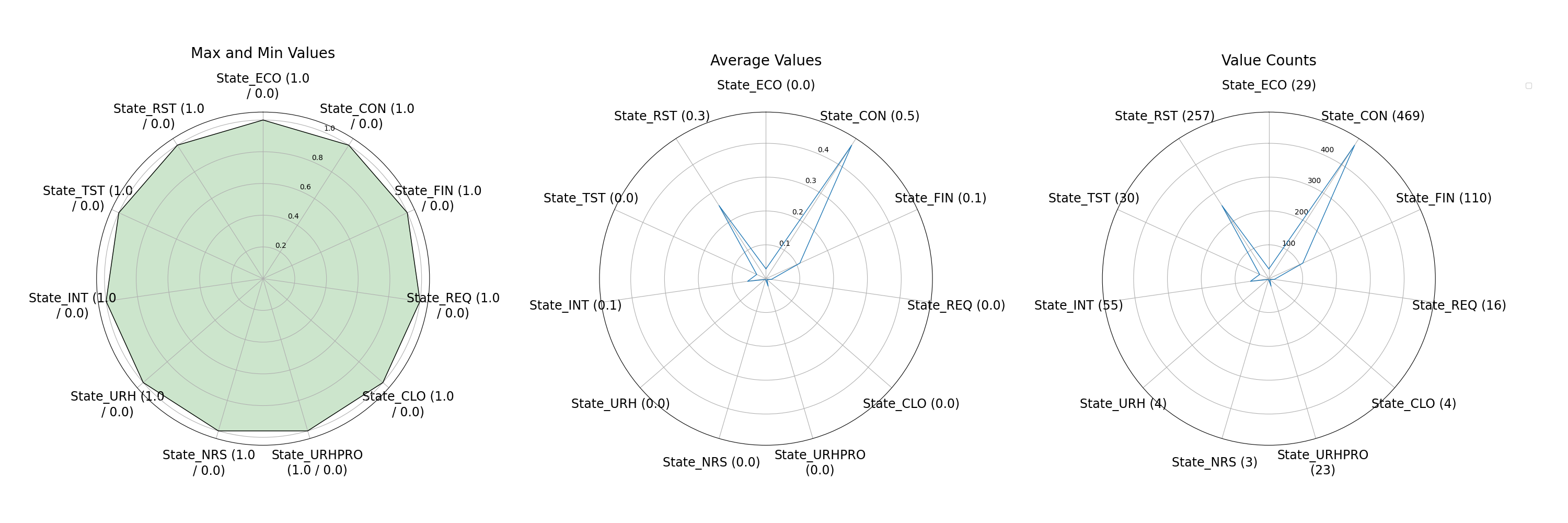}
    \caption{Resulting State category statistics for HopSkipJumpAttack with constraints.}
    \label{fig:state_hopskipjump_constrained}
    \end{figure}

    \begin{figure}[htpb]
    \centering
    \includegraphics[scale=0.15]{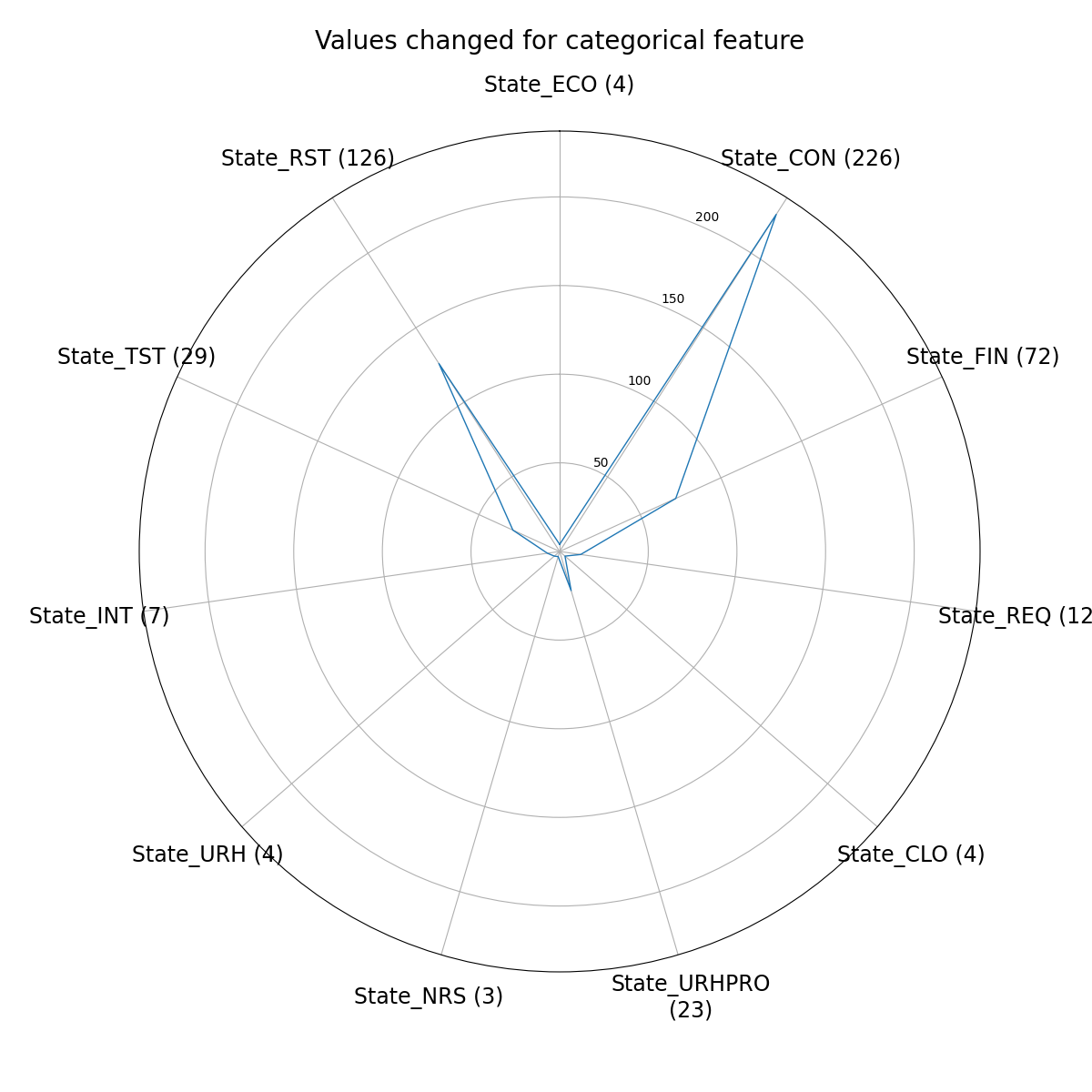}
    \caption{Changes in the State category for HopSkipJumpAttack with constraints.}
    \label{fig:state_hopskipjump_constrained_changes}
    \end{figure}
    
In contrast, the constrained variant of HopSkipJumpAttack produces adversarial examples that respect the logic of tabular data, as shown in Fig.~\ref{fig:state_hopskipjump_constrained}. The minimum and maximum values fall within the valid range of 0 and 1, the average values sum to 1, and the value counts sum to the number of successfully perturbed samples, with the value count distributions remaining consistent with one another.

As shown in Fig.~\ref{fig:state_hopskipjump_constrained_changes}, the change distribution for the constrained variant differs considerably from that of the unconstrained version. Beyond presenting lower overall values, consistent with the other figures, Fig.~\ref{fig:state_hopskipjump_constrained_changes} no longer closely mirrors the "Value Counts" distribution found in Fig.\ref{fig:state_hopskipjump_constrained}. While the clean samples had multiple columns with no values assigned, the adversarial examples now have values for every column, meaning that values from other, more populated columns, such as RST, were changed to other columns. Furthermore, RST was changed 126 times, although the original samples had 371, and the adversarial samples have 257. This suggests that, more than values changing from RST to others, other values also changed to RST.

Although both variants of HopSkipJumpAttack differ substantially in the quality of the adversarial samples produced, the unconstrained version achieves a higher MR of 81.8\%, compared to 42.9\% for the constrained untargeted variant. However, as demonstrated in Fig.~\ref{fig:state_hopskipjump}, the vast majority of adversarial samples produced by the unconstrained method are invalid and unrealistic. When evaluation treats low-quality samples on equal terms with high-quality ones, standard metrics may appear favorable; however, since these samples are, in most cases, both unrealistic and invalid, their impact on the overall robustness assessment should be weighted accordingly. By contrast, the constrained method, while unable to perturb every sample, produces valid and realistic adversarial examples. Taking this into account, the AURORA Distance Adjustment reduces the effective MR of the unconstrained variant to 0.12\%, while leaving the constrained variant unaffected.

\begin{figure}[htpb]
    \centering
    \includegraphics[width=\textwidth]{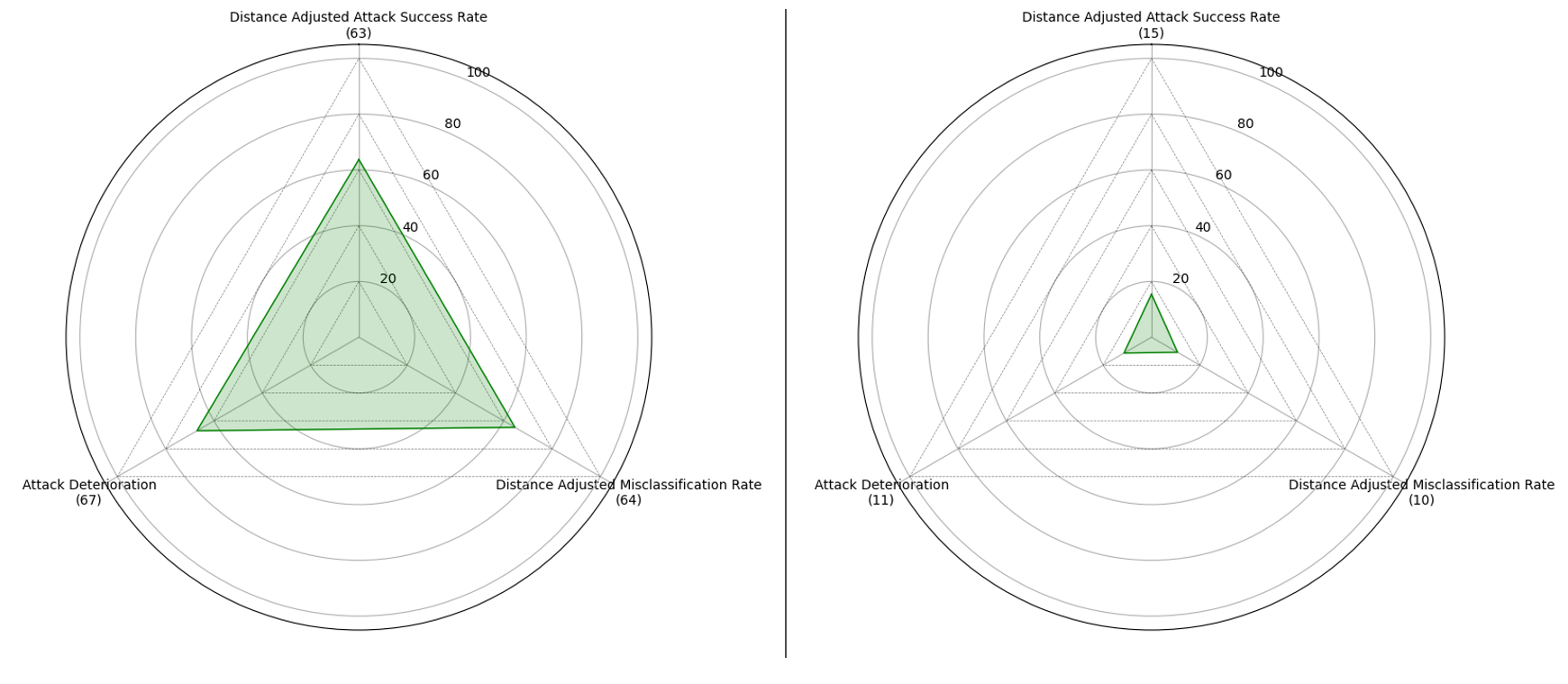}
    \caption{CatBoost robustness score on constrained perturbation methods. Standard evaluation case on the left, and worst-case scenario on the right.}
    \label{fig:robustness_cat}
    \end{figure}

As a final robustness evaluation of the CatBoost model, and based on the two previously introduced scenarios, the model achieves a robustness score of 65 when evaluated with methods fitted for tabular data, under the standard evaluation scenario, as shown in Fig.~\ref{fig:robustness_cat}, corresponding to a robust, near-moderately robust classification. While some perturbation methods successfully perturbed almost every clean sample, others failed to produce a single adversarial example, resulting in a more moderate average attack performance. Under the worst-case scenario, however, the outcome changes drastically, with the robustness score dropping to 13, corresponding to a not robust classification. This is attributable to high-performing constrained perturbation methods, which generate adversarial samples that are close to the original examples, respect the intrinsic constraints of tabular data, and therefore require no distance adjustment, yielding high metric values.

\begin{figure}[htpb]
    \centering
    \includegraphics[width=\textwidth]{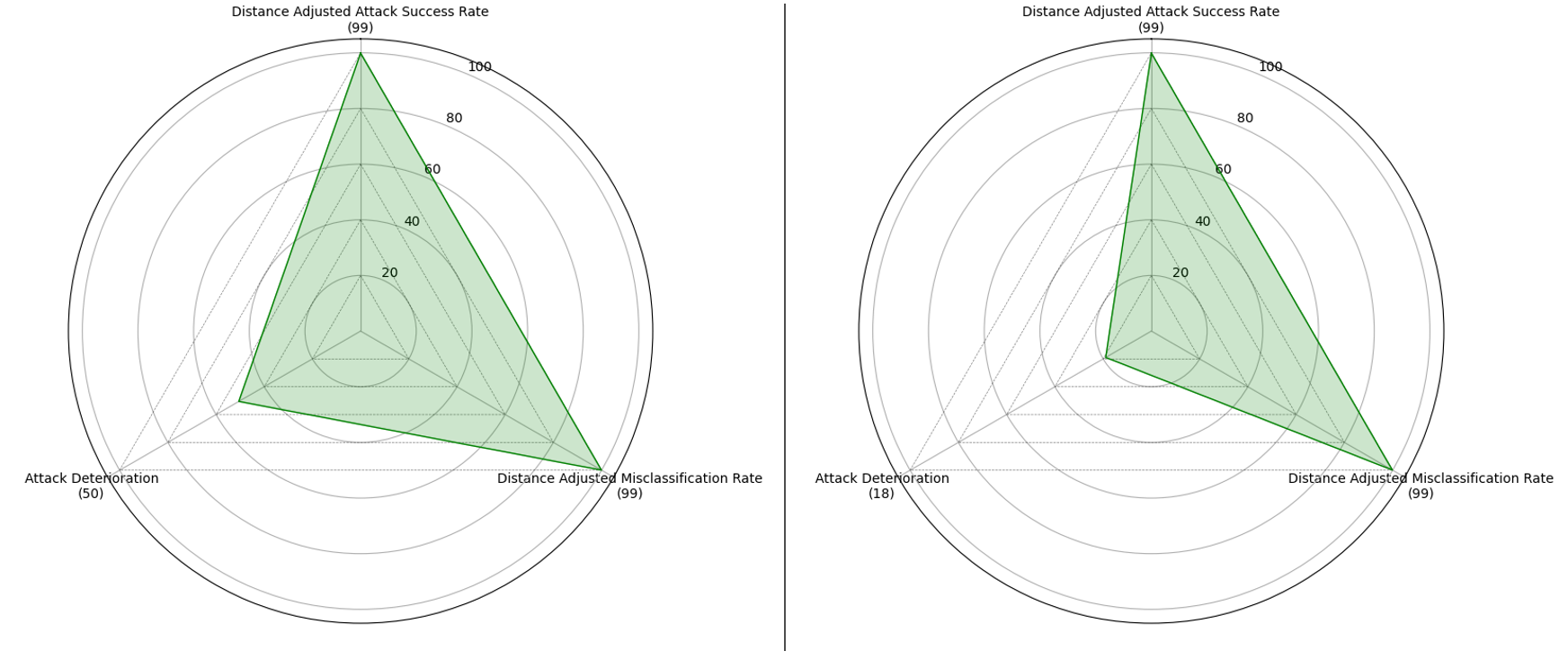}
    \caption{CatBoost robustness score on unconstrained perturbation methods. Standard evaluation case on the left, and worst-case scenario on the right.}
    \label{fig:robustness_cat_unconstrained}
    \end{figure}

The evaluation made using the original versions of the perturbation, however, differs significantly from the previous score. As the methods create unrealistic and invalid perturbations, the results are substantially discounted after the metric adjustment, further limiting their contribution to the final score. As such, the model achieves a robustness score of 83 under the standard evaluation scenario, as shown in Fig.~\ref{fig:robustness_cat_unconstrained}, a very robust score, while achieving 73 on the worst-case scenario, which translates to a robust score. It is important to note that this score should not be considered to be an absolute evaluation of the model robustness, since the model was tested using inappropriate perturbation methods, but as a reflection on how a user may misjudge a model as robust when testing it wrongly. This score should be taken into account, as well as the previously discussed figures, to ensure the safe deployment of an AI model.

\section{Conclusion}
\label{conclusion}
This paper presents AURORA, a framework for assessing the robustness of ML models in tabular data, addressing the critical lack of realism and validity in the adversarial samples typically used for such evaluations. AURORA streamlines the robustness evaluation process by integrating multiple perturbation methods and a diverse set of evaluation metrics, while being designed for scalability and adaptability across different scenarios. AURORA's novel distance adjustment mechanism treats categorical and numerical features separately, enabling more realistic and reliable robustness assessments by decoupling the final evaluation from results derived from low-quality adversarial examples.

To make this evaluation process transparent and actionable, AURORA compiles detailed information on each perturbation method, artifacts, and their results into a comprehensive report. Users can explore both the obtained metrics and the feature analysis, allowing them to understand the final robustness score and the reasoning behind it. This tool was designed to raise the standard for model robustness testing, enabling researchers and practitioners to easily evaluate their own ML models before deployment and select the most appropriate one for their specific use case through the two introduced evaluation scenarios.

Experiments demonstrated that conventional evaluation approaches may overestimate model vulnerability by treating unrealistic adversarial samples, which represent impossible real-world scenarios, on equal terms with valid ones. AURORA addresses this by incorporating distance-based penalties that distinguish valid perturbations from invalid ones, which leads to a more meaningful and trustworthy robustness evaluation. Rather than treating the final robustness score as a silver bullet, AURORA provides additional means to assess the quality of the data used in the evaluation, helping analysts understand the resulting robustness score and whether the testing methodology should incorporate different perturbation methods.

As most perturbation methods were originally developed for image data, future work will focus on developing perturbation strategies that are natively constrained to the valid feature space of each tabular data scenario. Given the strong results achieved by the constrained variants, a promising direction is to adapt these methods to enforce validity at each perturbation step, leveraging the distance metrics computed by AURORA to guide the process toward realistic examples, rather than transforming invalid adversarial samples into valid ones as a final step. Furthermore, including more objectives of attacks other than targeted or untargeted, as well as considering more constraints related to dependencies between features, could be pertinent. Additionally, testing the robustness of other model architectures on other datasets could provide insights on which perturbation methods are more fitted to tabular data.

\section*{Acknowledgment}

This research was supported by the SAFE project, which has received funding from the European Cybersecurity Competence Centre under grant agreement 101190370. Furthermore, the project UID/00760/2025 has also funded this work.

\bibliographystyle{plain}
\bibliography{bibliography}
\end{document}